# Detecting the Most Unusual Part of a Digital Image

Kostadin Koroutchev[1] and Elka Korutcheva[2] *

[1] EPS, Universidad Autónoma de Madrid,
Cantoblanco, Madrid, 28049, Spain
k.koroutchev@uam.es

[2] Depto. de Física Fundamental,
Universidad Nacional de Educación a Distancia,
c/ Senda del Rey 9, 28080 Madrid, Spain
elka@fisfun.uned.es

**Abstract.** The purpose of this paper is to introduce an algorithm that can detect the most unusual part of a digital image. The most unusual part of a given shape is defined as a part of the image that has the maximal distance to all non intersecting shapes with the same form.
The method can be used to scan image databases with no clear model of the interesting part or large image databases, as for example medical databases.

## 1 Introduction

In this paper we are trying to find the most unusual/rare part with predefined size of a given image. If we consider an one-dimensional quasi-periodical image, as for example electrocardiogram (ECG), the most unusual parts with length about one second will be the parts that correspond to rhythm abnormalities [6]. Therefore they are of some interest. Considering two dimensional images, we can suppose that the most unusual part of the image can correspond to something interesting of the image.

Of course, if we have a clear mathematical model of what the interesting part of the image can be, it would be probably better to build a mathematical model that detects those unusual characteristics of the image part that are interesting. However, as in the case of ECG, the part that we are looking for, can not be defined by a clear mathematical model, or just the model can not be available. In such cases the most unusual part can be an interesting instrument for screening images.

To state the problem, we need first of all a definition of the term "most unusual part". Let us chose some shape $S$ within the image $A$, that could contain that part and let us denote the cut of the figure $A$ with shape $S$ and origin $\boldsymbol{r}$ by $A_S(\boldsymbol{\rho}; \boldsymbol{r})$, e.g.

$$A_S(\boldsymbol{\rho}; \boldsymbol{r}) \equiv S(\boldsymbol{\rho})A(\boldsymbol{\rho} + \boldsymbol{r}),$$

where $\boldsymbol{\rho}$ is the in-shape coordinate vector, $\boldsymbol{r}$ is the origin of the cut $A_S$ and we used the characteristic function $S(.)$ of the shape $S$. Further in this paper we will omit the arguments of $A_S$. We can suppose that the rarest part is the one that has the largest distance with the rest of the cuts with the same shape.

Speaking mathematically, we can suppose that the most unusual part is located at the point $\boldsymbol{r}$, defined by:

$$\boldsymbol{r} = \arg\max_{\boldsymbol{r}} \min_{\boldsymbol{r}': |\boldsymbol{r}' - \boldsymbol{r}| > \text{diam}(S)} ||A_S(\boldsymbol{r}) - A_S(\boldsymbol{r}')||. \qquad (1)$$

Here we assume that the shifts do not cross the border of the image. The norm $||.||$ is assumed to be $L_2$ norm[1]

Because the parts of an image that intersect significantly are similar, we do not allow the shapes located at $r'$ and $r$ to intersect, avoiding this by the restriction on $r' : |\boldsymbol{r}' - \boldsymbol{r}| > \text{diam}(S)$.

If we are looking for the part of the image to be rare in a context of an image database, we can assume that further restrictions on $r'$ can be added, for example restricting the search to avoid intersection with several images.

The definition above can be interesting as a mathematical construction, but if we are looking for practical applications, it is too strict and does not correspond exactly to the intuitive notion of the interesting part as there can be several interesting parts. Therefore the correct definition will be to find the outliers of the distribution of the distances between the blocks $||.||$.

If the figure has $N^2$ points, and $||S|| \ll ||A||$, in order to find deterministically the most interesting part, we need $N^4$ operations. This is unacceptable even for large images, not concerning image databases. Therefore we are looking for an algorithm that provides an approximate solution of the problem and solves it within some probability limit.

As is defined above in Eq.(1), the problem is very similar to the problem of location of the nearest neighbor between the blocks. This problem has been studied in the literature, concerning Code Book and Fractal Compression [1]. However, the problem of finding $\boldsymbol{r}$ in the above equation, without specifying $\boldsymbol{r}'$, as we show in the present paper, can be solved by using probabilistic methods avoiding slow calculations.

Summarizing the above statements, we are looking for an algorithm that two blocks are similar or different with some probability.

---

[1] Similar results are achieved with $L_1$ norm. The algorithm was not tested with $L_{max}$ norm due to its extreme noise sensitivity. We use $L_2$ because of its relation with PSNR criteria that closely resembles the human subjective perception.

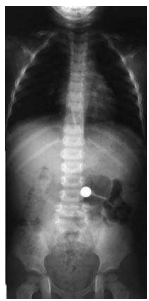
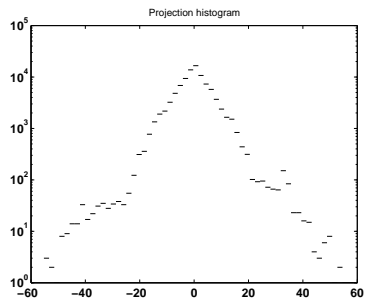

**Fig. 1.** The original test image. X-ray image of a person with ingested coin.

**Fig. 2.** The distribution of the projection value for square shape with a size 48x48 pixels.

## 2 The Method

### 2.1 Projections

The problem in estimating the minima of Eq. (1) is complicated because the block is multidimensional. Therefore we can try to simplify the problem by projecting the block $B \equiv A_S(\boldsymbol{r})$ in one dimension using some projection operator $X$. For this aim, we consider the following quantity:

$$b = |X.B_1 - X.B| = |X.(B_1 - B)|, \ |X| = 1. \tag{2}$$

The dot product in the above equation is the sum over all $\rho$-s:

$$X.B \equiv \sum_{\boldsymbol{\rho}} X(\boldsymbol{\rho}) B(\boldsymbol{\rho}; \boldsymbol{r}).$$

If $X$ is random, and uniformly distributed on the sphere of corresponding dimension, then the mean value of $b$ is proportional to $|B_1 - B|$; $\langle b \rangle = c|B_1 - B|$ and the coefficient $c$ depends only on the dimensionality of the block. However, when the dimension of the block increases, the two random vectors ($B_1 - B$ and $X$) are close to orthogonal and the typical projection is small. But if some block is far away from all the other blocks, then with some probability, the projection will be large. The method resembles that of Ref. [4] for finding nearest neighbor.

As mentioned above we ought to look for outliers in the distribution. This would be difficult in the case of many dimensions, but easier in the case of one dimensional projection.

We will regard only projections orthogonal to the vector with components proportional to $X_0(\rho) = 1, \forall \rho$. The projection on the direction of $X_0$ is proportional to the mean brightness of the area and thus can be considered as not so important characteristics of the image. An alternative interpretation of the above statement is by considering all blocks to differ only by their brightness.

Mathematically the projections orthogonal to $X_0$ have the property:

$$\sum_{\boldsymbol{\rho}} X(\boldsymbol{\rho}) = 0. \qquad (3)$$

The distribution of the values of the projections satisfying the property (3) is well known and universal [10] for the natural images. The same distribution seems to be valid for a vast majority of the images. The distribution of the projections derived for the X-ray image, shown in Fig. 1, is shown in Fig. 2.

Roughly speaking, the distribution satisfies a power law distribution in log-log scale if the blocks are small enough with exponential drop at the extremes. When the blocks are big enough, the exponential part is predominant.

If $A_r$ and $A'_r$ have similar projections, then they will belong to one and the same or to neighbors bins.

Therefore we can look for blocks that have a minimal number of similar and large projections. But these, due to the universality of the distribution, are exactly the blocks with large projection values.

As a first approximation, we can just consider the projections and score the points according to the bin they belongs to. The distribution can be described by only one parameter that, for convenience, can be chosen to be the standard deviation $\sigma_X$ of the distribution of $X.B$.

The notion of "large value of the projection" will be different for different projections but will be always proportional to the standard deviation.[2] Therefore we can define a parameter $a$ and score the blocks with $|X.B| > a\sigma_X$.

This procedure consists of the following steps:

0. Construct a figure $B$ with the same shape as $A$ and with all pixels equal to zero.

1. Generate a random projection operator $X$, with carrier with shape $S$, zero mean and norm one.

2. Project all blocks (convolute the figure). We denote the resulting figure as $C$.

3. Calculate the standard derivation $\sigma_X$ of the result of the convolution.

4. For all points of $C$ with absolute values greater than $a\sigma_X$, increment the corresponding pixel in B.

Repeat steps 1-4 for $M$ number of times.

5. Select the maximal values of $B$ as the most singular part of the image.

The number of iterations $M$ can be fixed empirically or until the changes in $B$, normalized by that number, become insignificant. Following the algorithm, one can see that the time to perform it is proportional to $MN^2 \log N$. The speed per image of size $1024 \times 2048$ on one and the same computer, with $S$, a square of size $56 \times 56$ points, is about 3 seconds compared to about an hour, using the direct search implementing the Eq. (1). [3]

---

[2] In general, the standard deviation will be larger for projections with larger low-frequency components. That is why we choose the criterion proportional to $\sigma_X$ and not as absolute value.

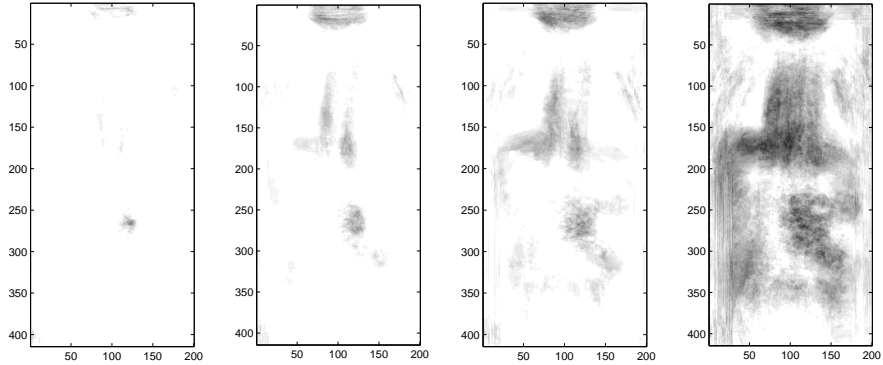

**Fig. 3.** Score values for different size of the shape (from left to right: 24x24,32x32,40x40,56x56). The value of the parameter $a$ in all the cases is 22.

Some results are presented in Figs.3,4, where we used square shapes with different size, 30 projection operators and different values of $a$.

Because the distribution of the projections (Fig. 2) is universal, it is not surprising that the algorithm is operational for different images. We have tested it with some 100 medical Xray images and the results of the visual inspections were good[4]

It can be noted that the number of projection operators is not critical and can be kept relatively low and independent of the size of the block. Note that with significantly large blocks, the results can not be regarded as en edge detector. This empirical observation is not a trivial result at all, indicating that the degrees of freedom are relatively few, even with large enough blocks, something that depends on the statistics of the images and can not be stated in general. With more than 20 projections we achieve satisfactory results, even for areas with more than 3000 pixels. The increment of the number of the projections improves the quality, but with more than 30 projection practically no improvement can be observed.

It is possible to look at that algorithm in a different way. Namely, if we are trying to reconstruct the figure by using some projection operators $X_C$ (for example DCT as in JPEG), then the length of the code, one uses to code a component with distribution like Fig. 2, will be proportional to the logarithm of the probability of some value of the projection $X_C.A$. Therefore, what we are scoring is the block that has some component of the code larger than some length in bits (here we ignore the psychometric aspects of the coding). Effectively we score the blocks with longer coding, e.g. the ones that have lower probability of occurrence.

---

[3] If the block is small enough, the convolution can be performed even faster in the space domain and it is possible to improve the execution time.

[4] Some of the images require normalization of the projection with the deviation of the block $B$.

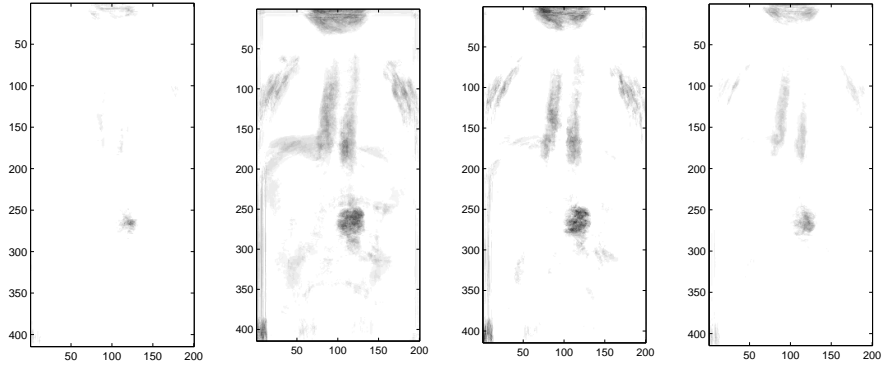

**Fig. 4.** Score values for different parameter $a$ (from left to right: $a = 8,10,12,16$). The size of the shape is 24x24.

Using a smoothed version of the above algorithm in step 4, without adding only one or zero, but for example, penalizing the point with the square of the projection difference in respect to the current block divided by $\sigma$, and having in mind the universal distribution of the projection, one can compute the penalty function as a function of the value of the projection $x$, that results to be just $1/2 + x^2/2\sigma^2$. Summing over all projections, we can find that the probability of finding the best block is approximated given by $1/2[1 + \text{erfc}(M(1/2 + x^2/2\sigma^2))]$ as a consequence of the Central Limit Theorem. The above estimation gives an idea why one need few projections to find the rarest block, in sense of the global distribution of the blocks, almost independently of the size of the block. The only dependence of the size of the blocks is given by $\sigma^2$ factor, that is proportional to its size. Further, the probability of error will drop better than exponentially with the increment of $M$.

The non-smoothed version performs somewhat better that the above estimation in the computer experiments.

### 2.2   Network

The pitfall of the consideration in the previous subsection is that the detected blocks are rare in absolute sense, e.g. in respect to all figures that satisfy the power law or similar distribution of the projections. Actually this is not desirable. If for example in X-ray image appear several spinal segments, although these can be rare in the context of all existing images, they are not rare in the context of thorax or chest X-ray images.

Therefore the parts of the images with many similar projections must "cancel" each other. This gives us the idea to build a network, where its components with similar projection are connected by a negative feedback corresponding to the blocks with similar projections.

As we have seen in the previous section, the small projection values are much more probable and therefore less informative. Using this empirical argument, we can suggest that the connections between the blocks with large projections are more significant.

The network is symmetrical by its nature, because of the reflexivity of the distances. We can try to build it in a way similar to the Hebb network [2] and define Lyapunov o energy function of the network. Thus the network can be described in terms of artificial recursive neural network. Connecting only the elements of the image that produce large projections, the network can be build extremely sparse [11], which makes it feasible in real cases.

Let us try to formalize the above considerations. For each point we define a neuron. The neurons corresponding to some point $r$ and having projection $x$ receive a positive input flux, which is proportional to $-\log p(x)$, where $p$ is the probability of having projection with value $x$. The same element, if its projection is large, also receives a negative flux from the points $r'$ with nearest projections that satisfy the condition $|r - r'| > \mathrm{diam}(S)$. The flux in general is a function of $p(x)$ and $x' - x$.

As a first approximation we assume that the flux is constant with $p(x)$ and the dependence on $x' - x$ is trivial: the weight is 1 if $|x' - x| < \delta$ and zero otherwise, where $\delta$ is some parameter of the model.

In other words, we reformulate our problem in terms of a Hebb-like neural network with external field

$$h = -h_0 \sum_{i=1}^{M} \log p(x_i) \qquad (4)$$

and weights

$$w_{rr'} = -\sum_{i=1}^{M} \sum_{\substack{|x_i| > a\sigma_i, \\ |x'_i| > a\sigma_i, \\ |x'_i - x_i| < \delta,\ x'_i x_i > 0}} 1. \qquad (5)$$

The extra parameter $h_0$ balances between the global and the local effects. It can be chosen in a way that the mean fluxes of positive and negative currents are equal in the whole network. The parameter $\delta$, as a proof of concept value, can be assumed to be equal to infinity. So the only parameter, as in the previous case, is $a$.

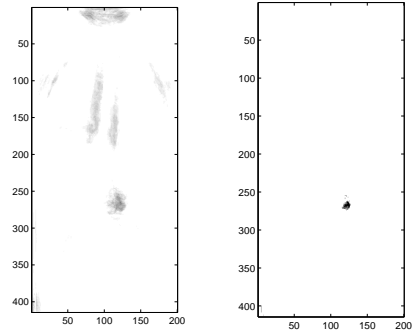

**Fig. 5.** Comparison between score image (left) and network activity image (right). The size of the area is 24x24 and the parameter $a = 16$.

The dynamics of the network over time $t$ is given by the following equation [3]:
$$s_{\bm{r}}(t+1) = g(\beta[h_r + \sum_{\bm{r'}} w_{\bm{rr'}} s_{\bm{r}}(t) - T]),$$
where $g(.)$ is a sigmoid function, $s_{\bm{r}}(t)$ is the state of the neuron $s$ at position $\bm{r}$ and time $t$, $\beta$ is the inverse temperature and $T$ is the threshold of the system. The result must be insensitive to the particular chose of $g(.)$.

Once the network is constructed, we need to choose its initial state. If the a priori probabilities for all points to be the origin of the rarest block are equal, one can choose $s_{\bm{r}}(0) = 1, \quad \forall \bm{r}$. Due to the non-linearity, the analysis of the results is not straightforward. The existence of the attractor is guaranteed by the symmetrical nature of the weights $w$, which is a necessary condition for the existence of an energy function.

We can further refine the results of the previous section by fixing the global threshold $T$ in a way to have only some fraction of the excited neurons. Thus we obtain a bump activity of the network, previously considered in [9,7,8]. A sample result is shown in Fig.5.

Regarding the time analysis of the procedure, one can see that the execution times are proportional to the number of the weights $w$. Having in mind that actually the connectivity is between the blocks, and that we can use a fraction of blocks less than $1/N^2$, the execution time can drop to order inferior to the $N^4$ limit. Thus, the number of steps to achieve the attractor is of the order $logN$.

## 3  Discussion and Future Directions

In this paper we present a method to find the most unusual (rare) part in two and higher dimensional images, when its shape is fixed, but in general arbitrary.

The method is almost independent on the size of the shape in terms of the execution speed and time. It gives good results on experimental images without predefined model of the interesting event.

One necessary future development of the algorithm is to achieve practical and computable criteria of the "rareness" of the block and comparing the results on large enough database in order to have qualitative measure of the results. The criterion must be different from Eq. (1), because its direct computing tends to be very slow and crispy.

Exact calculus of the probabilistic features of the network in the thermodynamics limit, performed in the sense of probability of finding the outliers, are also of common interest.

Among the future applications of the present method, one could mention the achievement of experiments on different type of images and large image databases and experiments on acceleration of the network due to the special equivalence class construction.

## 4  Acknowledgments

The authors acknowledge the financial support from the Spanish Grants TIN 2004–07676-G01-01, TIN 2007–66862 and DGI.M.CyT.FIS2005-1729 - Plan de Promoción de la Investigación UNED.

## References

1. Fisher Y., *Fractal Image Compression*, ISBN 0387942114, Springer Verlag (1995).
2. Hebb D., *The Organization of Behavior: A Neurophysiological Theory*, Wiley, New York (1949).
3. Hopfield J.:Neural networks and physical systems with emergent collective computational properties, Proc.Natl.Acad.Sci.USA, **79** (1982) 2554.
4. Indyk P.: Uncertainty Principles, Extractors, and Explicit Embedding of L2 into L1, in the Proceedings of 39th ACM Symposium on Theory of Computing, (2007).
5. Keogh E., Lin J. and Fu A.: HOT SAX: efficiently finding the most unusual time series subsequence, in the Proceedings of 39th ACM Symposium on Theory of Computing, (2007).
6. Keogh E., Lin in the Proceedings of the Fifth IEEE International Conference on Data Mining, (2005) 8.
7. Koroutchev K. and Korutcheva E.:Bump formations in binary attractor neural network, Phys.Rev.E **73**  (2006) 026107.
8. Koroutchev K. and Korutcheva E.: Bump formations in attractor neural network and their application to image reconstruction, in the Proceedings of the 9th Granada Seminar on Computational and Statistical Physics, AIP, ISBN 978-0-7354-0390-1 (2006) 242.
9. Roudi Y. and Treves A.: An associate network with spatially organized connectivity, JSTAT, **1** (2004) P07010.
10. Ruderman D.: The statistics of natural images, Network : Computation in Neural Systems, **5** (1994) 598.
11. Tsodyks M. and Feigel'man M.: The enhanced storage capacity in neural networks with low activity level, Europhys.Lett., **6** (1988) 101.

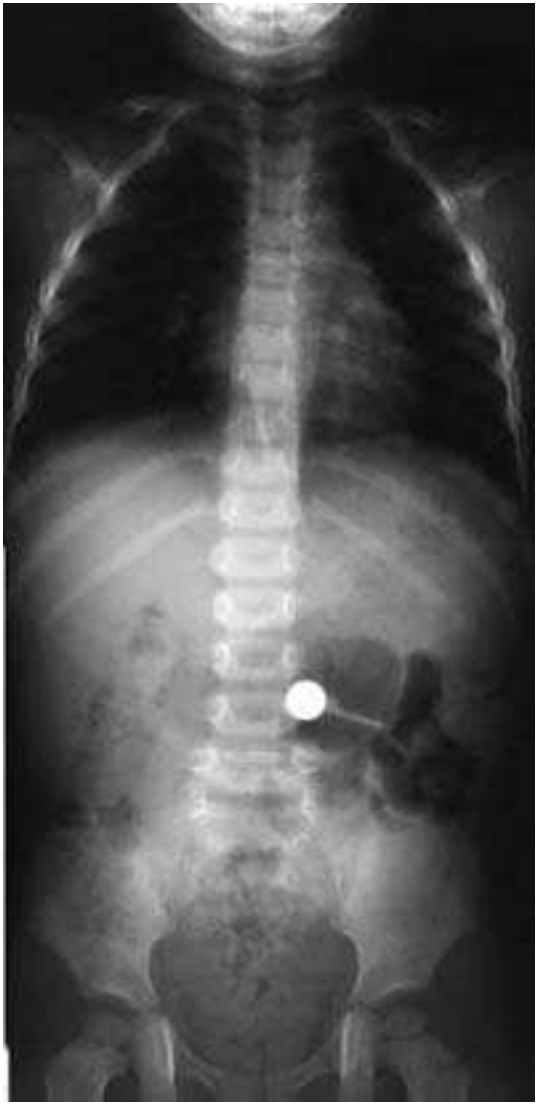

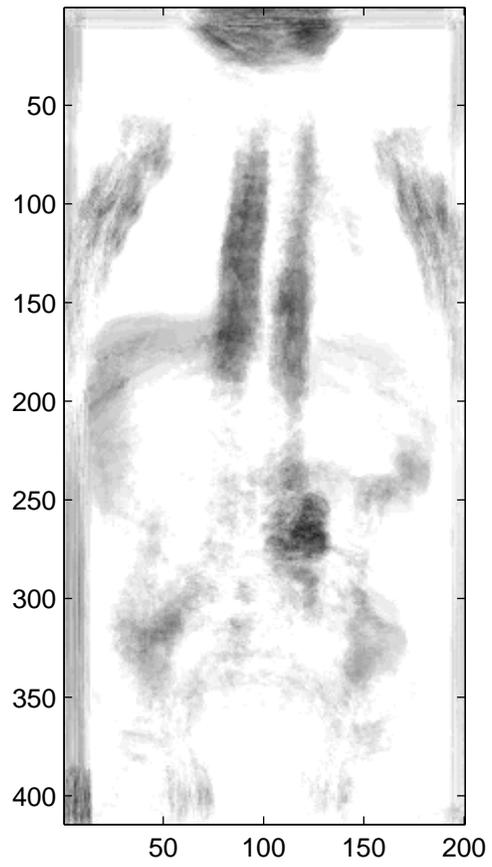

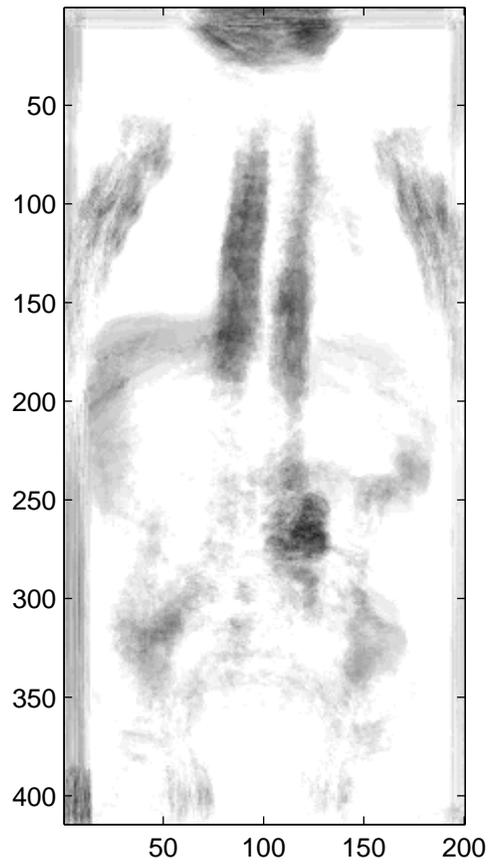

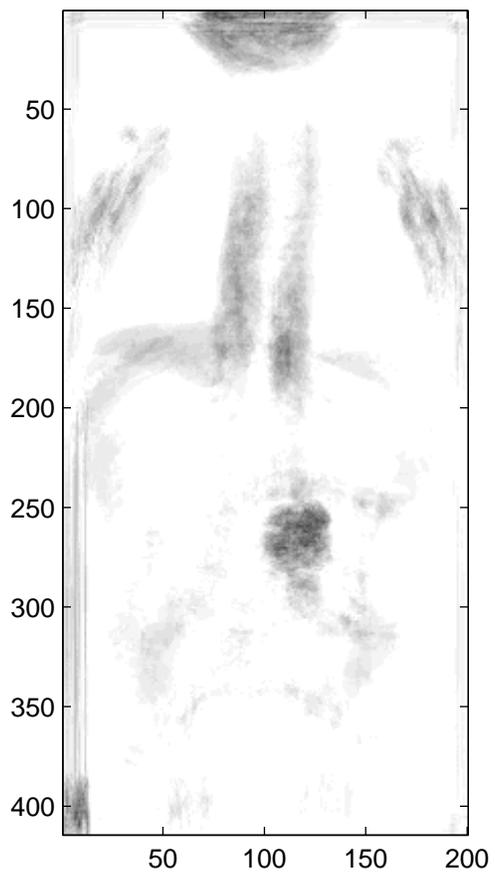

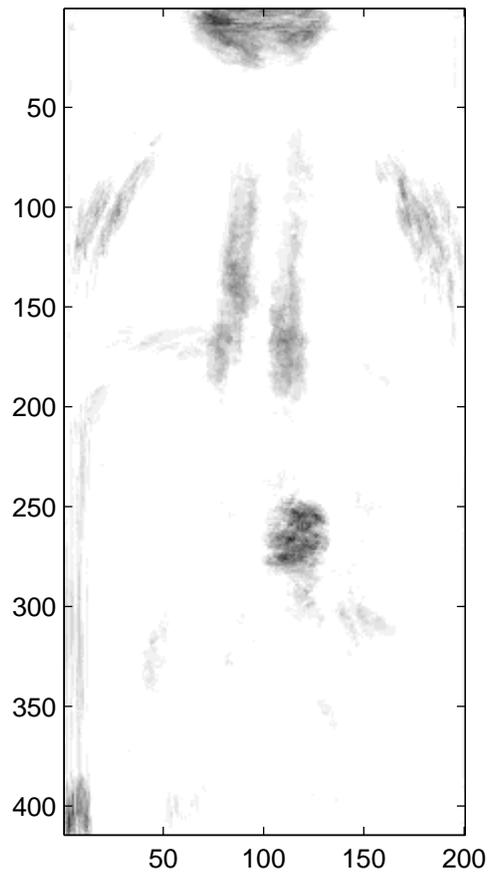

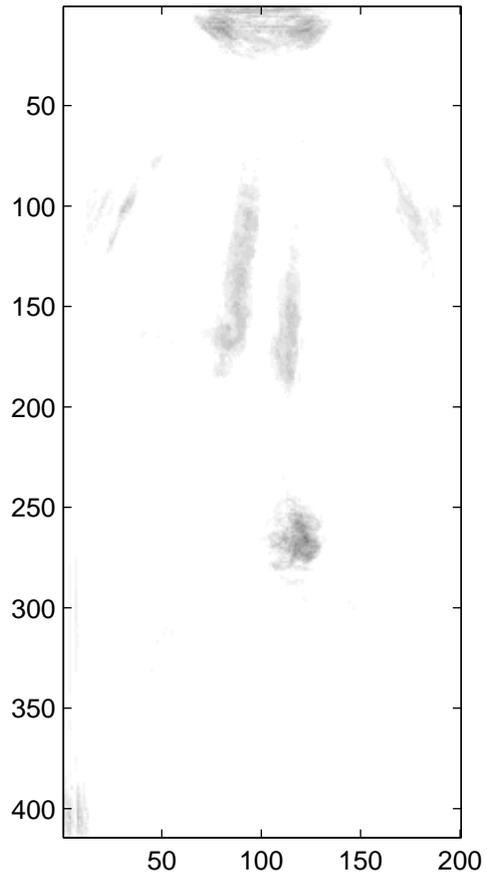

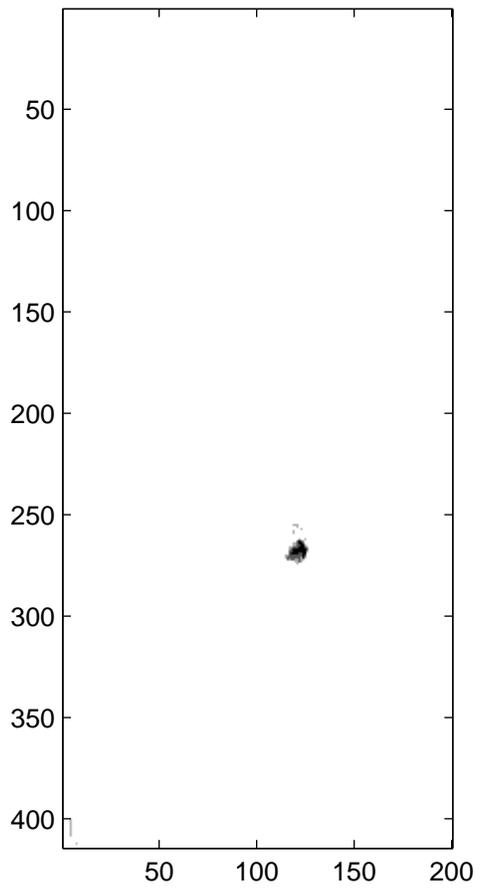

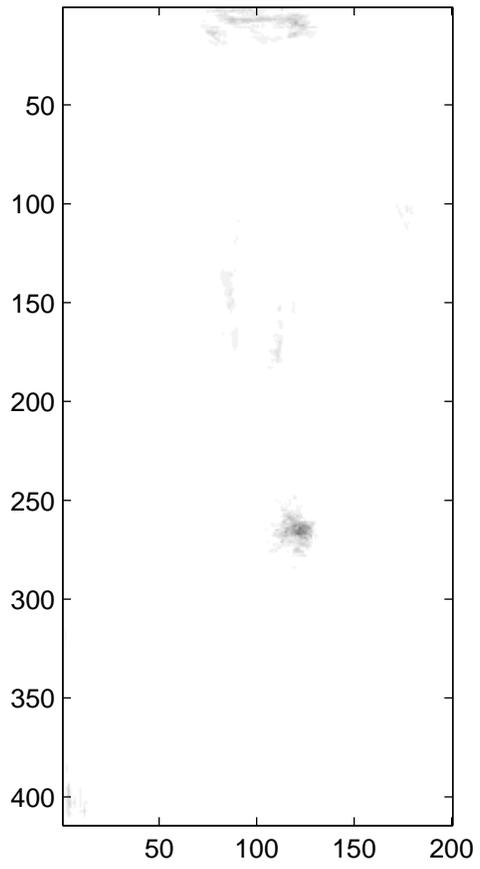

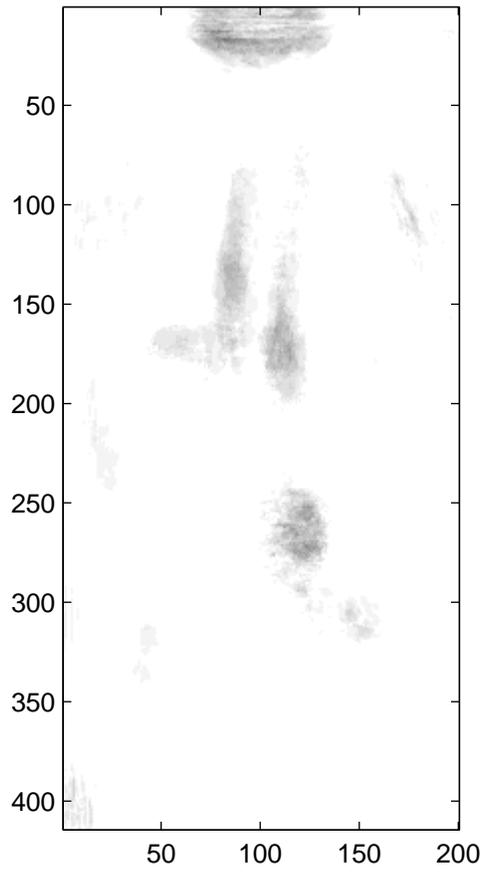

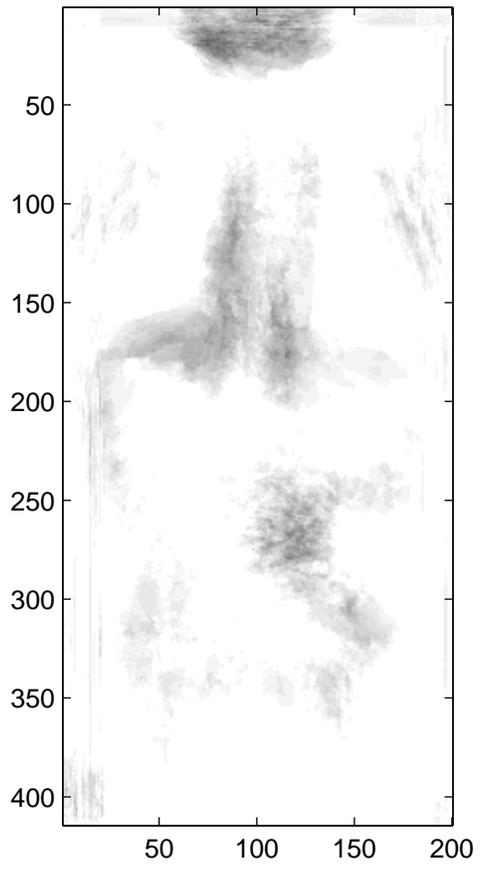

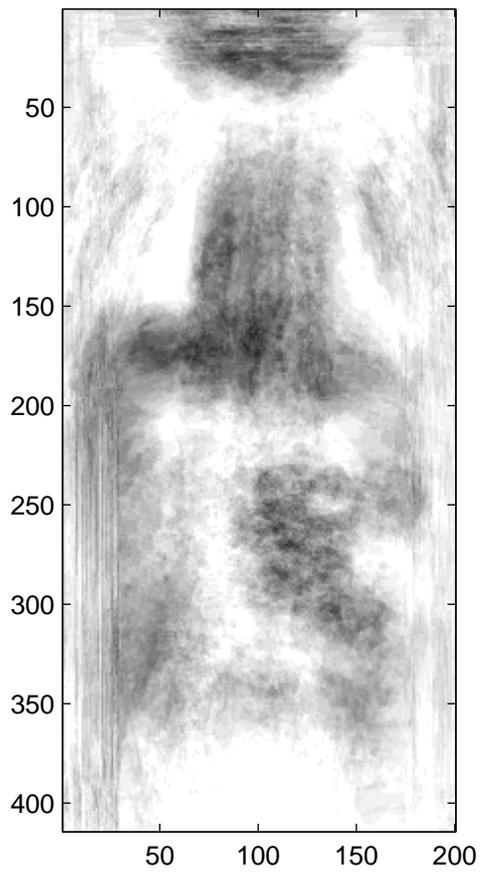

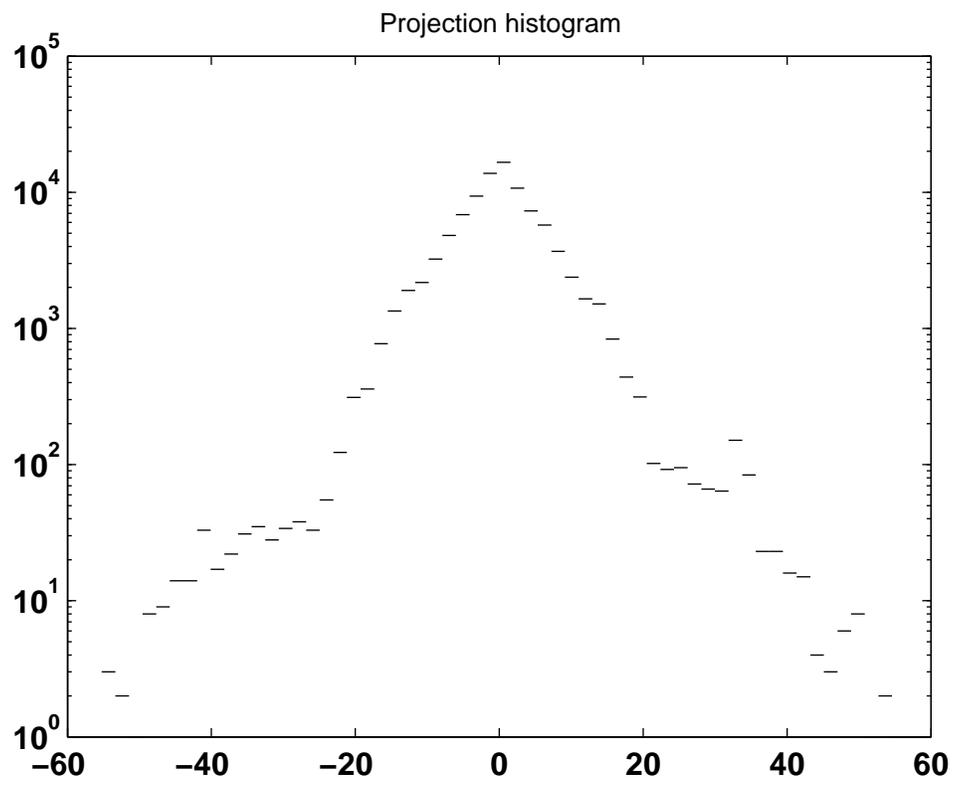

Projection histogram